\documentclass[10pt, twocolumn, letterpaper]{article}

\usepackage[utf8]{inputenc}
\usepackage[T1]{fontenc}
\usepackage{cuted}
\usepackage{algorithm}
\usepackage{algpseudocode}
\usepackage{newtxtext, newtxmath} 
\usepackage{geometry}
\usepackage{multirow}
\usepackage{makecell}
\usepackage{xcolor}
\usepackage{graphicx}
\usepackage{fancyhdr}
\usepackage{lipsum} 
\usepackage{tikz}   
\usepackage{titlesec} 
\usepackage{booktabs} 
\usepackage{authblk}  
\usepackage[hidelinks, colorlinks=true]{hyperref} 
\usepackage{eso-pic}  
\usepackage{dblfloatfix}

\usepackage[most]{tcolorbox}

\definecolor{OkestroBlue}{HTML}{0033A0} 
\definecolor{OkestroDark}{HTML}{001540} 
\definecolor{OkestroGrey}{HTML}{777777}
\definecolor{AbstractBG}{HTML}{E9E9E9}

\hypersetup{
    linkcolor=OkestroBlue,
    citecolor=OkestroBlue,
    urlcolor=OkestroBlue,
    pdftitle={Okestro Research Paper}
}

\AddToShipoutPictureBG*{%
    \begin{tikzpicture}[remember picture, overlay]
        \node[opacity=0.04] at (current page.south east) {
            \begin{tikzpicture}
                \fill[OkestroBlue] (-2,2) circle (12cm);
            \end{tikzpicture}
        };
        \node[opacity=0.03] at (current page.east) {
            \begin{tikzpicture}
                \fill[OkestroBlue] (0,5) circle (8cm);
            \end{tikzpicture}
        };
    \end{tikzpicture}
}

\titleformat{\section}
  {\sffamily\large\bfseries}
  {\thesection}{1em}{}
\titleformat{\subsection}
  {\sffamily\large\bfseries}
  {\thesubsection}{1em}{}

\newtcolorbox{abstractbox}{
  enhanced,
  colback=white,
  colframe=AbstractBG,
  arc=6pt,
  boxrule=0.6pt,
  left=8pt,
  right=8pt,
  top=8pt,
  bottom=8pt,
  width=0.88\textwidth
}

\title{
  \sffamily\bfseries
  Bridging Temporal and Textual Modalities: A Multimodal Framework for Automated Cloud Failure Root Cause Analysis
}

\author{
  {\large
    Gijun Park\textsuperscript{1,*}
  }\\
  {\small
    \textsuperscript{1}Okestro AI Research Center, Seoul, Republic of Korea
  }
}

\makeatletter

\footnotetext[1]{Corresponding author: Gijun Park (gj.park@okestro.com)}

\makeatother

\date{}

\begin{document}
\begin{strip}

\vspace{-5.0em}

\maketitle

\thispagestyle{fancy}

\begin{center}

\begin{abstractbox}
\small
\textbf{Abstract.}
Root cause analysis in modern cloud infrastructure demands sophisticated understanding of heterogeneous data sources, particularly time-series performance metrics that involve core failure signatures. While large language models demonstrate remarkable capabilities in textual reasoning, their discrete token-based architecture creates fundamental incompatibilities with continuous numerical sequences exhibiting temporal dependencies. Current methodologies inadequately address this modality mismatch, constraining the potential of language model-driven automation in incident management workflows. This paper presents a multimodal diagnostic framework that harmonizes time-series representations with pretrained language model embedding spaces. Our approach contributes three technical advances: (1) a semantic compression technique that distills temporal segments into single-token abstractions while preserving pattern semantics, (2) an alignment encoder utilizing gated cross-attention to project time-series features into language model latent space, and (3) a retrieval-augmented diagnostic pipeline that synthesizes aligned embeddings with historical incident knowledge for expert-level failure attribution. Comprehensive evaluation across six cloud system benchmarks demonstrates that our framework achieves leading performance, reaching 48.75\% diagnostic accuracy with notable improvements on scenarios involving compound failure modes. The results validate embedding-space alignment as an effective strategy for enabling language models to reason over multimodal telemetry data in production incident response contexts.

\vspace{1.2em}


\end{abstractbox}
\end{center}
\vspace{0.5cm}

\end{strip}

\section{Introduction}
\label{sec:intro}
In modern cloud computing environments, incident management and root cause analysis have become increasingly challenging due to the growing complexity of infrastructure and services. Concurrently, as digital transformation permeats mission-critical domains, site reliability engineering (SRE) which maintains system reliability and minimizes downtime through rapid incident identification and resolution has gained critical importance \cite{chigurupati_ai_2025,zurkowski_root_2024,li_fighting_2021}. To advance SRE, recruiting and retaining engineers with extensive incident investigation experience is crucial, yet this expertise is often scarce and costly to acquire, especially in new technology fields. Consequently, there is growing interest in automating incident investigation and diagnosis by leveraging the capabilities of large language models (LLM).

Incident investigation primarily involves analyzing various data sources, such as performance metrics, error logs, and trace logs, to derive and validate the root causes inferred from their characteristics \cite{lee_eadro_2023,yu_nezha_2023}. These data sources encompass multiple modalities, including the textual modality of error logs and the time-series modality of performance metrics. Among them, performance metrics contain temporal patterns and relationships that provide essential clues about the identity, scope, and timing of failures \cite{wu_identifying_2021,gu_identifying_2025}. Therefore, LLMs performing incident investigations are designed to derive root causes based on a comprehensive understanding of these data sources, knowledge of the cloud system and infrastructure that it can access externally or that is contained within its own parameters. Retrieval-augmented generation (RAG) enables this by integrating LLMs with external knowledge retrieval so that they can answer by referencing relevant information during inference \cite{lewis_retrieval-augmented_2020}. This approach enhances reasoning accuracy by retrieving relevant data from a vector store containing external knowledge sources, such as historical incident records and technical specifications based on semantic similarity between the query and stored embeddings \cite{zhang_cloudrca_2021, xu_openrca_2024}.

However, LLMs excel at processing discrete token sequences with semantic relationships inherent in textual data, they face challenges when applied to time-series data characterized by continuous numerical patterns and temporal dependencies. These temporal characteristics cannot be adequately captured through conventional tokenization methods \cite{10.5555/3692070.3692965,gruver_large_2023,niu2025langtime, tan2024are}. Moreover, representing numerical time-series data as text leads to exponential growth in token consumption, frequently exceeding LLM context length constraints \cite{10.5555/3692070.3692712,10.1145/3715073.3715083}. In addition, it is difficult to utilize external knowledge, most of which is composed of natural language, to aid  in domain knowledge and pattern understanding. These limitations become particularly acute for incident investigation with LLMs when analyzing non-textual data, such as time-series performance metrics. Despite the critical role of performance metrics in root cause analysis during incident investigations, their representation of time-series modality is not aligned with the text embedding spaces of LLMs \cite{10.1145/3691620.3695485,10.1145/3674726,10.1145/3629526.3645047}. This representational misalignment not only impedes automated incident diagnosis systems, but can also result in misinterpretation or overlooking crucial diagnostic indicators. The lack of integration of appropriate knowledge of the cloud domain or regard for data type further exacerbates these issues, leading to inconsistent and potentially erroneous diagnoses.

We propose TimeRAG\footnote{Our code will be made available upon acceptance.}, a novel framework that bridges the modality gap between time-series data and the pretrained embedding space of LLMs. TimeRAG aligns time-series performance metrics into the textual embedding space for LLMs to effectively understand and process. And it also incorporates empirical knowledge of cloud incident response via RAG, enabling expert-level diagnosis based on the integrated text-time-series multimodal queries.

The main contributions of this paper are as follows: (1) We introduce a single-token representation method that compresses time-series segments to only one semantic representation, preserving core temporal relationships with computational efficiency; (2) We propose a Time Series Encoder that aligns time-series patches with LLM embedding space through gated cross-attention mechanisms; (3) We develop an integrated RAG-based incident diagnosis system that combines aligned time-series embeddings with historical incident retrieval.

\section{Related Work}
Yu \textit{et al.} presented MonitorAssistant, an anomaly detection system for large-scale cloud services \cite{yu_monitorassistant_2024}. Their methodology automated model configuration recommendations by analyzing historical anomalies and generating structured, guidance-oriented reports using narratives generated by LLM. Zhang \textit{et al.} proposed an in-context learning approach leveraging GPT-4 for automated root cause analysis (RCA) \cite{zhang_automated_2024}. Rather than employing continuous fine-tuning, they used prompt engineering techniques to construct detailed and structured prompts, guiding the LLMs to generate accurate diagnostic outputs. Their findings highlighted the computational efficiency of context-driven prompt strategies for RCA tasks. Sarda \textit{et al.} introduced an LLM-powered on-call system that integrates multimodal datasets, including logs, metrics, traces, and alerts (LMTA), to automate RCA processes \cite{sarda_augmenting_2024}. Their approach aggregates runtime diagnostic information by capturing logs and metrics in real-time, correlating these with developer-defined alerts, and feeding the aggregated data into LLMs for root cause prediction. This method emphasizes the structuring of input information and its mapping to RCA categories.

Ahmed \textit{et al.} conducted an evaluation of LLMs for cloud incident management by examining multiple variants of GPT-3.x across zero-shot, low-rank adaptation (LoRA) fine-tuning, and multitask settings using a dataset of 40,000 incidents \cite{ahmed_recommending_2023}. They assessed performance through quantitative metrics and qualitative feedback from incident owners and demonstrated that fine-tuned LLMs ahead zero-shot approaches in effectively supporting incident resolution processes. Roy \textit{et al.} explored the use of a ReAct agent equipped with retrieval tools to enhance RCA capabilities \cite{roy_exploring_2024}. Their approach involved dynamic querying of diagnostic repositories and monitoring dashboards to supplement initial incident information. By iteratively refining its understanding of incidents, the agent demonstrated improved root cause identification accuracy. Las-Casas \textit{et al.} introduced LLexus, an AI agent system for automating troubleshooting guide (TSG) execution \cite{las-casas_llexus_2024}. Their method involved parsing TSGs into machine-readable steps, enabling the LLMs to autonomously follow procedural instructions and resolve incidents. Wang \textit{et al.} introduced RCAgent, an autonomous tool-augmented LLM framework. The system incorporated tools for real-time log analysis, data visualization, and interactive diagnostics \cite{wang_rcagent_2024}. By employing self-consistent action trajectories, RCAgent iteratively investigated incidents, significantly reducing manual efforts in RCA processes.

Despite this progress, existing LLM-based RCA approaches often fail to adequately integrate the temporal features of performance metrics. In addition, they frequently overlook the potential for deeper reasoning and hypothesis generation regarding causal relationships by not fully utilizing the inferential capabilities of LLMs.

\section{Methodology}

\begin{figure*}[t]
  \centering
  \includegraphics[width=\textwidth]{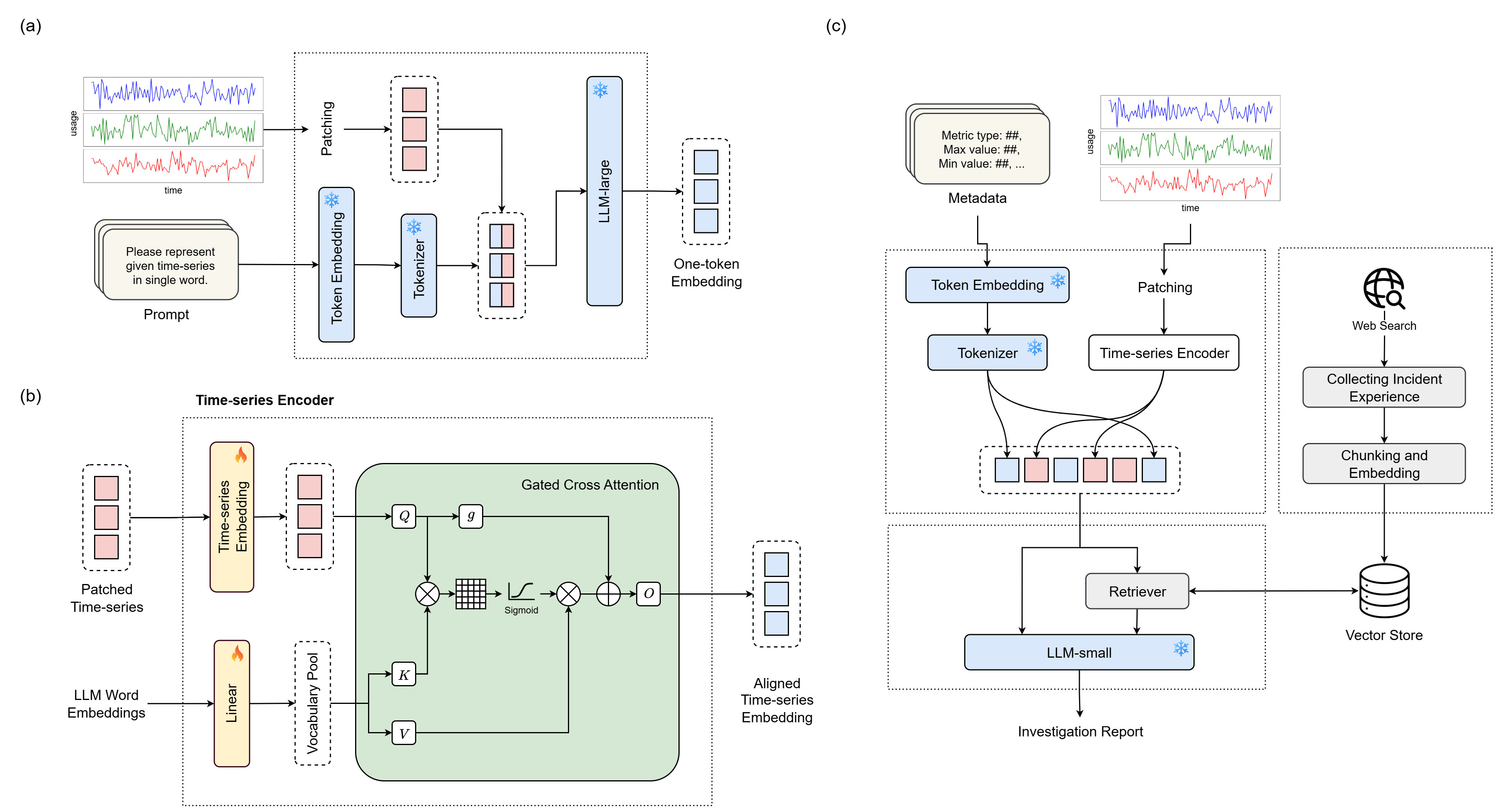}
  \caption{ TimeRAG overview. (a) time-series performance metrics are segmented into fixed-length patches, with each patch compressed into a single-token representation; (b) the Time Series Encoder maps the patch tokens into the pretrained LLM embedding space via gated cross-attention; and (c) the RAG diagnostic agent generates root-cause diagnoses based on aligned performance metrics and retrieved nearest incident embeddings from a vector store, with LLMs.}
\end{figure*}

TimeRAG enhances root cause analysis in cloud systems using LLMs with multimodal that understand time-series modalities of performance metric data. The methodology involves a series of stages, including generating semantic tokens projecting time-series patches, implementing encoders to align the meaning of temporal patterns to LLM embeddings using labeled time-series with the semantic token, and leveraging RAG-based incident retrieval with aligned performance metrics, metadata as a query, for cloud-specialized root cause diagnosis in the LLM.

\subsection{Patch Segmentation and Single-token Representation}

TimeRAG segments the time-series of performance metrics $X\in\mathbb{R}^{T\times F}$ into uniform patches $\{P_1,P_2,\ldots,P_N\}$ to create units that capture meaningful temporal relationships. These patches contain synchronized features of critical system metrics, including CPU usage patterns, memory consumption states, and GPU voltage variations \cite{nie_time_2022,tang_unlocking_2025}. However, since our goal is diagnosis rather than anomaly detection, we assume that the performance metrics are from periods already identified as containing a system failure.

Based on these patches, Patch Abstraction LLM, a model dedicated to extracting the meaning of patches, generates a single token that most closely matches the meaning of the given patch. For instance, a patch representing a stable trend would be generated to a token such as 'stable' or 'steady' by the model. These tokens that semantically project the patches become labels for those patches, compressing their numeric values.

\subsection{Time Series Encoder}

A Time Series Encoder is designed to identify time-series patterns in performance metric data and align these identified meanings from the time-series modality to the textual modality, harmonizing them within the embedding space of the LLM. Thereby enabling the pretrained LLM to directly understand the temporal and contextual information of the patches in the embedding space it already possesses. In addition, it ensures seamless integration with prompts from other text modalities, including metadata, within the same embedding space. Based on the Time-LLM architecture proposed by Ming \textit{et al.}, The encoder employs a modified two-stage alignment process: first, downsampling the LLM vocabulary to create a focused vocabulary pool to represent time-series data, then aligning time-series patterns using gated cross-attention mechanisms by mapping them into this vocabulary pool \cite{jin_time-llm_2023}.

First, the LLM vocabulary $\mathcal{V}_{\mathrm{LLM}}$ is downsampled to a smaller, trainable parameter $\mathcal{V}_{\mathrm{time}}$ that is tailored specifically for time-series data, as shown in equation 1. The vocabulary used to express time-series is limited, constituting less than one percent of the total LLM vocabulary. Thus, the vocabulary pool built through this downsampling process contains only core and meaningful terms. This reduction not only enhances computational efficiency, but also improves the ability to identify time-series patterns for the encoder.

\begin{equation}
  \mathcal{V}_{time}\subset\mathcal{V}_{LLM},\ \ \left|\mathcal{V}_{time}\right|\ll\left|\mathcal{V}_{LLM}\right|\
\end{equation}

The encoder segments the input performance metrics into patches and embeds them into the hidden dimension of the LLM. Then, each patch undergoes gated cross-attention with the vocabulary pool. During this process, the gate mechanism balances the degree of reflection between the embedding values and the attention outputs through the gate values, which are trainable parameters of the sigmoid function.

Performance metrics rarely show dynamic variation across the entire length of a sample, so usually more than half of the patches within a sample have the same single-token representation. Conventional cross-attention mechanisms may ignore subtle expression differences in these patch sequences, mapping them all uniformly to popular representations. In contrast, the gated cross-attention prevents this by dynamically modulating the attention of individual time-series patches based on the degree of agreement between mapped patches and the label. This approach mitigates the risk of uniform mapping and enhances the encoder's sensitivity to subtle deviations, which may be indicative of potential system failures.

\begin{algorithm}
\caption{TimeseriesEncoder}
\begin{algorithmic}[1]
\Require $\mathit{target} \in \mathbb{R}^{B \times L \times d_{\text{model}}}$, $\mathit{source}, \mathit{value} \in \mathbb{R}^{S \times d_{\text{llm}}}$
\State \textbf{Parameters:} Linear maps $q_{\text{proj}}, k_{\text{proj}}, v_{\text{proj}}, \text{out}_{\text{proj}}$, scalars $\text{temperature}, \text{gate}, \lambda_{\text{init}}$, vectors $\lambda_{q1}, \lambda_{k1}, \lambda_{q2}, \lambda_{k2}$, RMSNorm, $n_{\text{heads}}$, with $d_{\text{keys}} \approx d_{\text{model}}/n_{\text{heads}}$
\State $Q \leftarrow \text{reshape}\left(q_{\text{proj}}\left(\mathit{target}\right), \left(B, L, n_{\text{heads}}, d_{\text{keys}}\right)\right)$
\State $K \leftarrow \text{reshape}\left(k_{\text{proj}}\left(\mathit{source}\right), \left(S, n_{\text{heads}}, d_{\text{keys}}\right)\right)$
\State $V \leftarrow \text{reshape}\left(v_{\text{proj}}\left(\mathit{value}\right), \left(S, n_{\text{heads}}, d_{\text{keys}}\right)\right)$
\State $\text{scale} \leftarrow \text{temperature}/\sqrt{d_{\text{keys}}}$
\State $A \leftarrow \text{softmax}\left(\text{scale} \cdot \left(Q \cdot K\right), \text{ over source positions}\right)$
\State $\lambda_{\text{full}} \leftarrow \exp\left(\sum\left(\lambda_{q1} \odot \lambda_{k1}\right)\right) - \exp\left(\sum\left(\lambda_{q2} \odot \lambda_{k2}\right)\right) + \lambda_{\text{init}}$
\State $A \leftarrow \left(1 - \lambda_{\text{full}}\right) \cdot A$
\State $R \leftarrow \text{RMSNorm}\left(A \cdot V\right) \cdot \left(1 - \lambda_{\text{init}}\right)$
\State $O \leftarrow \text{gate} \cdot R + \left(1 - \text{gate}\right) \cdot \mathit{target}$
\State \Return $\text{out}_{\text{proj}}\left(\text{flatten}(O)\right)$
\end{algorithmic}
\end{algorithm}

In training, Time Series Encoder outputs embedding values for each patch and the predicted root cause types, from those embedding values by a classifier of a single fully-connected layer. Accordingly, the training objective is to minimize two losses: the cross-entropy between tokens derived from the embeddings of each patch and its corresponding label token, and the categorical cross-entropy between the predicted and true root causes. The tokens of patch embedding are obtained by a greedy search for logits from the output embedding layer. The combined training objective $\mathcal{L}$ can be formulated as follows:

\begin{equation}
  \mathcal{L} = \mathcal{L}_{\text{align}} + \mathcal{L}_{\text{clf}}
\end{equation}
The alignment loss $\mathcal{L}_{\text{align}}$ maps time-series embeddings to the LLM embedding space:

\begin{equation}
  \mathcal{L}_{\text{align}} = \text{CrossEntropy}(\mathbf{E}_{\text{LLM}}(h_{\text{align}}), \mathbf{y}_{\text{tokens}})
\end{equation}

where $h_{\text{align}}$ is the aligned representation of Time Series Encoder, $\mathbf{E}_{\text{LLM}}$ denotes the tokens derived from the LLM output embedding, and $\mathbf{y}_{\text{tokens}}$ represents the tokenized time-series targets. This loss encourages the model to learn representations that can be decoded into meaningful tokens representing temporal patterns.

The classification loss $\mathcal{L}_{\text{clf}}$ considers differences in the aligned representations by failure types.

\begin{equation}
  \mathcal{L}_{\text{clf}} = \text{CrossEntropy}(h_{\text{clf}}, \mathbf{y}_{\text{class}})
\end{equation}

where $h_{\text{clf}}$ is the output of the classification head and $\mathbf{y}_{\text{class}}$ denotes the failure class labels.

The encoder implements adaptive attention weighting with learnable parameters $\lambda_q$ and $\lambda_k$, modulating the alignment between time-series patches and LLM embeddings. During training, we implement dynamic token masking after the first epoch, probabilistically masking frequently predicted tokens to encourage diverse alignment patterns. The model is optimized using Adam with cosine annealing, while keeping LLM parameters frozen to preserve pretrained knowledge.

\begin{table*}[t]
  \centering
  \caption{Dataset Overview}
  \vspace{0.1cm}
  \begin{tabular}{cccc}
  \toprule
    \textbf{Dataset}&\textbf{Feature metrics}&\textbf{\makecell{Metric lengths\\/ Frequency}}&\textbf{Failure types}\\
    \midrule
    SockShop&\makecell{CPU, Memory, Network bandwidth\\(byte, packets)}&361 / 1s&\multirow{2}{*}{\makecell{CPU hog, Network delay,\\Packet loss, Memory leak}}\\
    OnlineBoutique&CPU, Memory, Workload, Latency&900 / 1s&\\
    Exathlon&CPU, Memory, Network bandwidth (byte)&900 / 1s&\makecell{Bursty/stalled input, CPU\\contention, Process failure}\\
    LemmaRCA(Prod)&\multirow{3}{*}{\makecell{CPU, Memory, Network bandwidth\\(byte, packets)}}&900 / 1s&Noisy neighbor, DDoS\\
    LemmaRCA(Cloud)&&900 / 1s&\makecell{SW bugs, Noisy neighbor,\\Malware infected, Resource limit}\\
    AIOpsArena&\makecell{CPU, Memory, Istio request and response\\(byte, mesessages)}&360 / 15s&\makecell{CPU stress, Pod failure,\\Packet loss, Network delay}\\
  \bottomrule
  \end{tabular}
\end{table*}

\subsection{Integreted RAG Agent}

The integrated RAG agent leverages aligned embeddings from the Time Series Encoder within an RAG framework to perform cloud-specialized root cause diagnosis. The agent consists of three primary components: (1) a vector store containing historical incident data, (2) a query processor that integrates aligned time-series embeddings with system metadata, and (3) a Diagnostic LLM that generates root cause analyses based on retrieved context.

The vector store contains documents of historical incident reports, resolution processes, and embedded representations of them. The documents are loaded into the vector store through the following pipeline: First, raw material is segmented into semantic chunks that preserve contextual integrity while maintaining manageable sizes for retrieval. These chunks typically contain problem descriptions, system states at failure time, diagnostic steps taken, and ultimate resolutions. Second, each chunk is embedded using the embedding model same of LLM processing the query to ensure compatibility with the query embeddings. Finally, the embedded chunks are indexed in the vector store along with their metadata.

The documents are constructed in a total of 3,249 cases with one document for each case, from diverse incident archives, including those from cloud service platforms, LLM-based chat services, and code configuration management systems. They are initially segmented into chunks of 512 tokens. A filtering process using Diagnostic LLM as a binary classifier retains only chunks containing incident symptoms and the corresponding resolutions, resulting in 4,796 chunks. This filtering removes administrative content and unrelated documentation.

The query processor composes into a coherent diagnostic query format by combining the aligned performance metrics embeddings and system metadata with a narrative template. This system metadata includes performance metric types, periods, maximum values, and minimum values. The system metadata provides performance metrics numerical range and a baseline for the temporal patterns, enabling LLM to accurately assume the original time-series even if most of the detailed numerical values are lost during the alignment process. When this diagnostic query is submitted, the agent performs similarity-based retrieval to identify the most relevant historical incidents. The retrieval process computes the cosine similarity between the embedding of the diagnostic query and all chunks in the vector store. The top-k most similar chunks are extracted, where k is a hyperparameter empirically set to 5 to balance context richness with computational efficiency.

The retrieved chunks, along with the original query, are then passed to a Diagnostic LLM to generate a root cause analysis. Diagnostic LLM generates a structured diagnostic report that identifies potential root causes and recommended remediation steps, synthesizing all of the inputs.

To enhance diagnostic accuracy, the agent implements a reflection mechanism that iteratively refines the initial diagnosis \cite{ji-etal-2023-towards}. After generating the initial root cause analysis, the Diagnostic LLM performs a self-evaluation to assess the completeness and consistency of its findings. This evaluation considers factors such as whether all anomalous patterns in the time-series have been addressed, whether the proposed root causes align with the retrieved historical cases, and whether the recommended actions are feasible given the system constraints.

If self-evaluation identifies deficiencies, the system triggers a refinement cycle. During refinement, the agent may request additional chunks from the vector store, specifically targeting aspects identified as incomplete in the evaluation. These new chunks replace some of the initially retrieved chunks based on their relevance to the identified gaps. The Diagnostic LLM then regenerates its analysis, incorporating both the evaluation feedback and the updated contextual information. This reflection process can iterate up to a predefined maximum of 5 or until the self-evaluation meets quality thresholds.

\section{Experiments}

\begin{table*}[t]
  \centering
  \caption{Experimental results of cloud incident root cause diagnosis with short-term performance metric}
  \vspace{0.1cm}
  \begin{tabular}{ccccc}
  \toprule
    \multirow{2}{*}{\makecell{\textbf{Models}\\}} & \multicolumn{2}{c}{\textbf{SockShop}} & \multicolumn{2}{c}{\textbf{AIOpsArena}} \\
    & \textbf{\textit{Acc.}} & \textbf{\textit{Win Rate}} & \textbf{\textit{Acc.}} & \textbf{\textit{Win Rate}} \\
    \midrule
    DeepSeek-R1-Distill-SRE-Qwen-32B-INT8&\textbf{28.89}&86.67&20.41&87.76\\
    kubeguru-llama3.2-3b-v0.1&17.78&100.00&14.29&95.92\\
    Mistral-7B-TimeSeriesReasoner&17.78&71.11&\textbf{26.53}&51.22\\
    ChatTS-14B&\underline{22.22}&53.33&20.41&63.27\\
    TimeRAG (Ours)&19.79&-&\underline{25.00}&-\\
  \bottomrule
\end{tabular}
\end{table*}

\begin{table*}[t]
  \centering
  \caption{Experimental results of  cloud incident root cause diagnosis with long-term performance metric}
  \vspace{0.1cm}
  \begin{tabular}{ccccccccccccc}
  \toprule
    \multirow{2}{*}{\makecell{\textbf{Models}\\}} & \multicolumn{2}{c}{\textbf{\makecell{Online\\Boutique}}} & \multicolumn{2}{c}{\textbf{Exathlon}} & \multicolumn{2}{c}{\textbf{\makecell{LemmaRCA\\(Prod)}}} & \multicolumn{2}{c}{\textbf{\makecell{LemmaRCA\\(Cloud)}}} \\
    & \textbf{\textit{Acc.}} & \textbf{\textit{Win Rate}} & \textbf{\textit{Acc.}} & \textbf{\textit{Win Rate}} & \textbf{\textit{Acc.}} & \textbf{\textit{Win Rate}} & \textbf{\textit{Acc.}} & \textbf{\textit{Win Rate}} \\
    \midrule
    \makecell{DeepSeek-R1-Distill-SRE-Qwen-32B-INT8}&\underline{25.00}&93.18&\underline{24.72}&97.75&11.92&98.01&\underline{30.60}&98.18\\
    kubeguru-llama3.2-3b-v0.1&22.73&100.00&15.73&100.00&\underline{27.81}&99.34&18.58&100.00\\
    Mistral-7B-TimeSeriesReasoner&11.36&63.64&23.60&70.79&21.19&78.15&15.12&83.97\\
    ChatTS-14B&22.73&52.27&21.35&53.93&13.91&55.63&15.85&58.29\\
    TimeRAG (Ours)&\textbf{43.75}&-&\textbf{28.91}&-&\textbf{48.75}&-&\textbf{31.60}&-\\
  \bottomrule
\end{tabular}
\end{table*}

\subsection{Experimental Setup}

\subsubsection{Datasets}
TimeRAG is evaluated on 6 datasets: online-butique, sock-shop, exathlon, aiops-arena, and lemma-rca \cite{jacob_exathlon_2021,sun_scenario-oriented_2024,pham_baro_2024,pham_rcaeval_2025,zheng_lemma-rca_2025}. The exathlon dataset comprises performance metrics collected during failure events in Apache Spark clusters. The remaining datasets contain performance metrics obtained through simulated direct and indirect failure scenarios for e-commerce service workloads. For model implementation and validation, each dataset was partitioned with an 80-20 split for training and test, respectively. The training dataset is used for training the Time-series Encoder, and the test dataset is used to validate the performance of the entire TimeRAG framework. Table 1 presents detailed characteristics of each dataset, including sample sizes, feature types, and failure types.

\subsubsection{Preprocessing}
Except for lemma-rca and exathlon, which are constant at 900, the other datasets do not have a consistent length of the performance metric in each sample. When comparing the length of the longest samples for each dataset, there is a large difference, with a minimum of 361 and a maximum of more than 50,000. Meanwhile, the length per sample of the exathlon dataset is constant at 900. Therefore, considering that many cloud failures are transient or brief in nature, we adopt 900 time steps, equivalent to 15 minutes, which is reasonably extended from the typical cases as the standardized length, and standardize the samples included in datasets other than lemma-rca and exathlon according to this criterion \cite{285740,10.1186/s13677-018-0112-9}. First, samples exceeding 900 time steps were patched using a non-overlapping sliding window. Subsequently, samples shorter than 900 time steps were augmented to match the standardized length using Time-LLM, a time-series prediction model. The model was trained on if the maximum sample length of each dataset is shorter than 900, all samples of max length; otherwise all samples of length 900. Lastly, the samples that need to be augmented are extrapolated by forecasting from the end of the sample to the length of the training data.

\subsubsection{Baselines}
We evaluate TimeRAG against 4 baseline approaches: (1) DeepSeek-R1-Distill-SRE-Qwen-32B-INT8, which optimized specifically for operations and SRE scenarios; (2) Kubeguru-Llama3.2-3B, specialized providing guidance and question-answering for Kubernetes operations, Linux administration, and containerized application management; (3) Mistral-7B-TimeSeriesReasoner, fine-tuning LLM designed to analyze server performance metrics and predict potential failures; and (4) ChatTS-14B, multimodal LLM, enabling it to perform both understanding and reasoning with time-series, treating time-series and text as modalities \cite{noauthor_phpcooldeepseek-r1-distill-sre-qwen-32b-int8_2024,noauthor_spectro-cloudkubeguru-llama32-3b-v01_nodate,noauthor_esperantomistral-7b-timeseriesreasoner_2024,xie_chatts_2025}. For the Patch Abstraction LLM, we use Qwen3-235B-A22B \cite{yang_qwen3_2025}. Meanwhile, the Diagnostic LLM, which is integrated into the RAG engine, employs Deepseek-R1-Distill-Qwen-7B. \cite{deepseek-ai_deepseek-r1_2025}. All baseline experiments utilize identical input formats, fixed seed (0), and hardware configurations (4 NVIDIA A100 40GB GPUs) to ensure fair comparative analysis. 

\subsubsection{Evaluation Metric}
The performance of incident investigation and root cause diagnosis are evaluated using accuracy, which represents the percentage of correct answers selected for multiple-choice questions consisting of 4 for lemma-rca or 5 for others choices from the benchmark dataset. For the win rate, both the baseline model and TimeRAG generate incident reports based on performance metrics and their metadata in given failure situations, unlike the accuracy. GPT-4o then evaluates which of the two generated reports is more accurate and comprehensive, and the win rate represents the percentage of cases where the proposed model outperforms the baseline.

\subsection{Results}

In long-term performance metrics (900 time steps), TimeRAG achieves accuracies ranging from 28.91\% to 48.75\%. Specifically, LemmaRCA (Prod) yields the highest accuracy at 48.75\%, where the dataset contains complex failure scenarios including noisy neighbors, DDoS attacks, and malware infections. Online Boutique, which includes various failure types such as CPU stress, pod failures, and network delays, results in 43.75\% accuracy.

For shorter time sequences (360-361 time steps), TimeRAG achieves 19.79\% accuracy on SockShop and 25.00\% on AIOps Arena. This presents different challenges due to their abbreviated temporal windows, which contain less contextual information for pattern recognition.

The experimental results indicate that TimeRAG's alignment mechanism between time-series embeddings and LLM space enables effective root cause analysis across various failure scenarios. TimeRAG processes both simple failures (such as CPU hog and memory leak) and complex, multifaceted incidents (including distributed denial of service and resource contention) with measurable accuracy levels suitable for practical deployment in cloud incident diagnosis systems.

\section{Conclusion}
We proposed TimeRAG, a novel framework that bridges the representational gap between time-series performance metrics and pretrained LLM embedding spaces for cloud incident diagnosis. TimeRAG is configured with a single-token representation, Time Series Encoder, and an integrated RAG agent.

Experimental evaluation in 6 RCA benchmark datasets of cloud system failures demonstrates the effectiveness of TimeRAG, achieving an SOTA accuracy of up to 48.75\%. The framework shows particular strength on complex failure scenarios in LemmaRCA datasets, validating that explicit alignment between time-series and textual modalities significantly enhances diagnostic capabilities. These results establish TimeRAG as a practical solution for automated root cause analysis in cloud environments, offering an approach to leverage pretrained LLMs for temporal pattern understanding without extensive retraining.

TimeRAG can substantially reduce mean time to resolution in production environments by automating first-response diagnosis from an incident declaration. This capability makes the system organize by providing augmented decision support to junior engineers. Furthermore, by providing broader access to advanced diagnostic capabilities, TimeRAG can help improve service reliability and reduce operational overhead across SRE teams.

\section{Limitations and Future Work}
TimeRAG demonstrates promising results, several limitations warrant consideration. The preprocessing pipeline's augmentation and windowing strategies may introduce artifacts that affect diagnosis accuracy. In serious failure scenarios that last several hours or more, the limited numerical information provided by the system metadata has limitations in ensuring the numerical accuracy and resolution of aligned performance metrics that LLMs understand. TimeRAG was evaluated on GPU-equipped servers in offline settings, leaving real-time deployment constraints unexplored. Furthermore, the performance of RAG depends on the quality and coverage of historical incidents in the vector store, which potentially limits the effectiveness of novel failures.

Future work will evaluate robustness to noisy telemetry data and data from production environments, while also introducing a human-in-the-loop validation process with SRE practitioners to assess operational impact. Beyond that, we will explore extending TimeRAG to real-time streaming scenarios in a production environment, and adapting the alignment methodology to other multimodal analysis tasks farther on cloud incident diagnosis for more generalization.

\renewcommand{\refname}{\normalsize References}

\begingroup
\small
\bibliographystyle{IEEEtran}
\bibliography{references}
\endgroup

\end{document}